\documentclass[runningheads]{llncs}

\usepackage{eccv}

\usepackage{eccvabbrv}

\usepackage{graphicx}
\usepackage{booktabs}
\usepackage[table]{xcolor}
\usepackage{float}

\usepackage[accsupp]{axessibility}  

\usepackage{hyperref}

\usepackage{orcidlink}

\newcommand{\best}[1]{\textcolor{red}{#1}}
\newcommand{\second}[1]{\textcolor{blue}{#1}}
\newcommand{\venue}[1]{{\color{gray}\tiny(#1)}}
\newcommand{\figsub}[2]{Fig.~\hyperref[#1]{\ref*{#1}#2}}

\begin{document}


\title{CoCo-SAM3: Harnessing Concept Conflict in Open-Vocabulary Semantic Segmentation}

\titlerunning{CoCo-SAM3 for Open-Vocabulary Semantic Segmentation}

\author{Yanhui Chen\textsuperscript{1} \and
Baoyao Yang\textsuperscript{1} \and
Siqi Liu\textsuperscript{2} \and
Jingchao Wang\textsuperscript{3}}

\authorrunning{Y.~Chen et al.}

\institute{\textsuperscript{1} School of Computers, Guangdong University of Technology \quad
\textsuperscript{2} Shenzhen Research Institute of Big Data \quad
\textsuperscript{3} Peking University\\
\email{chenyanhui91@mails.gdut.edu.cn, ybaoyao@gdut.edu.cn, siqiliu@sribd.cn, ethanwangjc@163.com}}

\maketitle

\begin{abstract}
SAM3 advances open-vocabulary semantic segmentation by introducing a prompt-driven mask generation paradigm. However, in multi-class open-vocabulary scenarios, masks generated independently from different category prompts lack a unified and inter-class comparable evidence scale, often resulting in overlapping coverage and unstable competition. Moreover, synonymous expressions of the same concept tend to activate inconsistent semantic and spatial evidence, leading to intra-class drift that exacerbates inter-class conflicts and compromises overall inference stability. To address these issues, we propose \textbf{CoCo-SAM3} (Concept-Conflict SAM3), which explicitly decouples inference into intra-class enhancement and inter-class competition. Our method first aligns and aggregates evidence from synonymous prompts to strengthen concept consistency. It then performs inter-class competition on a unified comparable scale, enabling direct pixel-wise comparisons among all candidate classes. This mechanism stabilizes multi-class inference and effectively mitigates inter-class conflicts. Without requiring any additional training, \textbf{CoCo-SAM3} achieves consistent improvements across eight open-vocabulary semantic segmentation benchmarks.

\end{abstract}

\section{Introduction}
\label{sec:intro}

\noindent Open-vocabulary semantic segmentation (OVSS)~\cite{bucher2019zero,xian2019semantic} aims to assign semantic labels to image pixels without requiring a pre-fixed category set. Unlike conventional semantic segmentation, which relies on a closed label system and assumes the same categories during training and testing, OVSS models categories as open concepts described in natural language. This eliminates the need for costly and repeated annotation and retraining when new categories emerge, significantly reducing deployment overhead and enabling scalable adaptation to novel concepts. As such, OVSS provides essential capabilities for open-world understanding and interactive vision applications.

\begin{figure}[!t]
  \centering
  \IfFileExists{fig/fig1.pdf}{%
    \includegraphics[width=\linewidth]{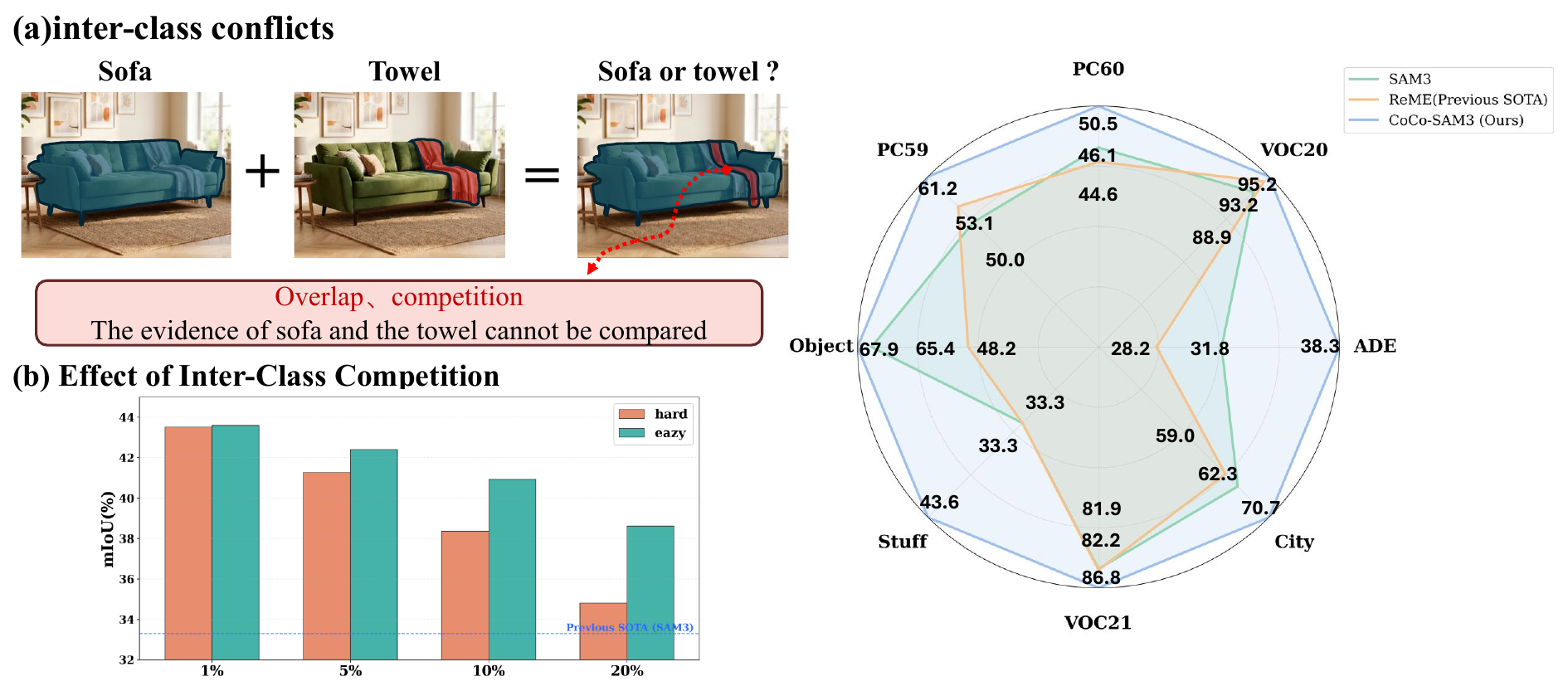}%
  }{%
    \fbox{\rule{0pt}{1.6in}\rule{0.98\linewidth}{0pt}}%
  }
    \caption{
    Left: Inter-class conflicts when applying SAM3 to open-vocabulary semantic segmentation. Masks for different categories are generated independently from their respective prompts, without evidence calibration on a unified scale, resulting in mutual overwriting and confusion. Right: Controlled inter-class competition analysis on COCO-S. We vary the ratio of inter-class competitors $p$ in the semantic-prior normalization by adding to the target class the most similar ``easy'' and the least similar ``hard'' negative classes (non-target classes), and report mIoU.
    }
  \label{fig:intro}
\end{figure}

Recent advances in OVSS have increasingly focused on training-free inference, which leverages pre-trained foundation models to segment novel concepts directly without task-specific fine-tuning. This paradigm circumvents the annotation bottleneck and enables rapid deployment to evolving category sets, aligning with the core motivation of OVSS. Within this paradigm,  mainstream approaches typically treat CLIP~\cite{radford2021learning} as a source of semantic evidence, producing pixel-level category responses by measuring the similarity between dense visual features and text embeddings. Along this line, ClearCLIP~\cite{lan2024clearclip} improves the usability of open-vocabulary prediction mainly by eliminating irrelevant structures and achieving more stable vision--language alignment, while SFP enhances spatial coherence by adaptively detecting and suppressing the propagation of abnormal token responses in attention. However, relying solely on CLIP-based semantic responses often fails to provide reliable pixel-accurate boundaries and fine-grained structures, thereby limiting contour quality and localization precision. To compensate for missing structural cues, another family of training-free methods incorporates additional vision foundation models (\textit{e.g.}, DINOv2~\cite{oquab2023dinov2} and SAM2~\cite{ravi2024sam}) as priors for structure and boundary separability (\textit{e.g.}, FreeDA\cite{barsellotti2024training} and CorrCLIP~\cite{zhang2025corrclip}). Although such multi-model collaboration can improve contours and details, the resulting pipelines and cross-representation alignment significantly increase inference complexity; more importantly, these methods still lack mechanism-level constraints on cross-class interactions in multi-class dense prediction, and thus cannot fundamentally suppress the instability induced by inter-class conflicts.

 
As a new generation of promptable segmentation foundation model, SAM3~\cite{carion2025sam} further advances OVSS from similarity-driven discriminative prediction to a concept-conditioned paradigm: conditioned on a concept prompt, it directly generates a segmentation mask for the corresponding concept, providing a more expressive and scalable pathway for OVSS~\cite{kirillov2023segment}.
However, since SAM3 is trained with one-to-one supervision between a single word and its mask, its outputs are inherently on a prompt-conditioned scale. When directly applied to multi-concept joint inference, mask responses triggered by different category prompts lack a unified, inter-class comparable evidence scale, making it difficult to establish stable relative priorities and mutual exclusion across concepts; this leads to regional overwriting and inter-class conflicts (see the overlapping competition between ``sofa'' and ``towel'' in \figsub{fig:intro}{a}). Moreover, we conduct a controlled inter-class competition analysis on the COCO-S dataset by varying the ratio of negative classes $p$ used in semantic-prior normalization, and comparing performance when selecting the most similar ``easy'' versus the least similar ``hard'' negative classes as competitors. The results in \figsub{fig:intro}{b} show that both the strength and composition of inter-class competition significantly affect mIoU, indicating that multi-class OVSS requires explicit and controllable modeling and calibration of inter-class competition, rather than allowing incomparable prompt responses to passively conflict in space.

Moreover, we find that in multi-class joint inference, the instability of inter-class conflicts can be largely attributed to insufficient intra-class evidence consistency. Since SAM3's responses operate on a prompt-conditioned scale, the same semantic concept expressed with different phrasings (including synonyms) may activate inconsistent semantic facets and spatial evidence, causing intra-class evidence to scatter or even contradict each other. This lack of intra-class coherence undermines a concept's stable representativeness in inter-class competition, thereby amplifying uncertainty in cross-class competition and making overwriting and conflicts more likely. Consequently, before establishing stable inter-class competition, we must consolidate multi-view intra-class evidence into a consistent representation.

Motivated by these observations, this paper proposes \textbf{CoCo-SAM3}, a training-free framework that stably transfers SAM3's promptable mask generation capability to open-vocabulary multi-class semantic segmentation. \textbf{CoCo-SAM3} keeps SAM3 parameters frozen and constructs an intermediate representation with stronger semantic--text alignment on dense features from an intermediate layer of the perception encoder, serving as a unified interface for multi-concept joint inference. This representation provides two core functions. On the one hand, it establishes global cross-class calibration that maps pixel-level evidence of different concepts onto a unified, inter-class comparable scale, such that competition among semantically similar concepts at the same spatial location attains stable relative priorities and mutual exclusivity; this imposes mechanism-level suppression of inter-class conflicts and thereby substantially alleviates spatial overlap, overwriting, and confusing assignments. On the other hand, synonym prompts are treated as multi-view expressions of the same concept, and synonym evidence is calibrated and weight-aggregated only within this representation; under the same unified scale, intra-class semantic cues are integrated to improve intra-class semantic consistency and reduce sensitivity to prompt wording variations. Under these comparable and consistent semantic constraints, the concept masks generated by SAM3 under prompt conditioning are fused as structural evidence, so that SAM3's strong contour priors support stable multi-class semantic decisions. The main contributions are summarized as follows:

\begin{itemize}
    \item From the perspective of multi-class open-vocabulary joint inference, we identify two critical limitations in SAM3: the absence of a unified evidence scale for cross-concept competition, and the lack of intra-class semantic consistency maintenance under open-naming diversity. We attribute the resulting systemic instability to SAM3's output organization and its prompt-conditioned response mechanism.
    \item We propose \textbf{CoCo-SAM3}, a training-free framework that constructs an alignment-enhanced intermediate representation within the perception encoder. It establishes a unified scale and stably models inter-class competition through global cross-class calibration, while simultaneously calibrating and aggregating evidence from synonymous prompts within this representation to restore intra-class semantic consistency.
    \item Extensive experiments on eight OVSS benchmarks demonstrate that \textbf{CoCo-SAM3} consistently improves multi-class joint inference performance under the training-free setting. It significantly reduces conflicts among semantically similar concepts and mitigates fluctuations caused by synonym prompts, validating its effectiveness in stabilizing open-vocabulary segmentation.
\end{itemize}



\section{Related Work}
\label{sec:related}

Open-Vocabulary Semantic Segmentation (OVSS) aims to perform pixel-level semantic segmentation under an open label space, conditioned on category concepts described in natural language. Unlike conventional semantic segmentation that relies on a fixed label set with matched training and test categories, OVSS allows the category set to be specified dynamically at inference time and requires the model to generalize to unseen categories and diverse textual expressions.

\noindent{\normalsize\textbf{Train-based OVSS.}} Train-based methods typically rely on pixel-level mask annotations or text supervision for learning~\cite{ghiasi2022scaling,liang2023open,ding2022decoupling,xie2024sed,xu2023masqclip, shi2026mmerror}. For example, LSeg~\cite{li2022language} directly learns pixel-level representations and aligns them with textual concepts, enabling segmentation with text categories as queries; CAT-Seg~\cite{cho2024cat} further formulates pixel-to-text matching as a cost volume and performs aggregation to enhance fine-grained structures and local discrimination. Beyond fully supervised settings, weakly supervised approaches such as GroupViT~\cite{xu2022groupvit} explore grouping mechanisms that allow semantic region structures to emerge from text supervision, offering an alternative way to reduce annotation dependence. Overall, train-based methods often achieve a higher performance ceiling, but they require supervision and training pipelines, and typically need retraining or additional adaptation when transferring across domains or when the label space changes.

\noindent{\normalsize\textbf{Training-free OVSS: CLIP-only.}} This line of work treats CLIP’s aligned representations as the sole semantic source and converts them into pixel-level category responses through inference-time structural modifications and response calibration~\cite{bousselham2024grounding,hajimiri2025pay,zhou2022extract}. Many studies generalize the query–key attention in CLIP’s last layer to self–self attention variants or their combinations to reduce noise in dense matching and improve localization, such as the value–value attention in CLIP Surgery~\cite{li2023clip}, the query–query and key–key attention in SCLIP~\cite{wang2024sclip}, and the more general self–self attention combinations in GEM~\cite{bousselham2024grounding}. Other works diagnose and adjust CLIP’s final-layer architecture to improve dense separability; for instance, ClearCLIP~\cite{lan2024clearclip} processes the last-layer residual branch and feed-forward structure to obtain cleaner segmentation responses. Overall, these methods are lightweight and easy to deploy, but their pixel-accurate boundaries and fine-grained structural modeling are often bounded by CLIP’s image–text alignment pretraining objective.

\noindent{\normalsize\textbf{Training-free OVSS: CLIP-VFM.}} 
Another direction introduces vision foundation models as structural priors to constrain, reorganize, or fuse CLIP’s dense responses, compensating for its limitations in localization and boundary delineation~\cite{lan2024proxyclip,kim2025distilling,barsellotti2024training,shi2025harnessing,zhang2025corrclip,lu2026chordedit}. FreeDA~\cite{barsellotti2024training} leverages DINOv2~\cite{oquab2023dinov2} self-supervised representations to enhance local separability and structural awareness; ProxyCLIP~\cite{lan2024proxyclip} constructs more reliable pixel-level proxy representations via external visual priors to improve spatial consistency in open-vocabulary semantic segmentation; CorrCLIP~\cite{zhang2025corrclip} combines DINO~\cite{caron2021emerging} and SAM2~\cite{ravi2024sam} priors to reconstruct and constrain CLIP patch correlations, improving contour quality and fine-grained structures. In general, this collaborative line can substantially enhance boundaries and region structures, but it also introduces additional complexity due to multi-model pipelines and cross-representation alignment. Moreover, it still lacks direct inference-time constraints for cross-class interference in multi-class open-vocabulary settings and within-class inconsistency caused by naming diversity, and thus may remain unstable under large sets of semantically similar categories or varying synonym prompts.

\section{Preliminaries: SAM3}

\noindent\textbf{Architecture.} SAM3~\cite{carion2025sam} is a promptable foundation segmentation model that formulates segmentation as \emph{concept-conditioned mask generation}: given an input image \(I\) and a concept prompt \(s\) (\textit{e.g.}, a class name or a short phrase), the model outputs a binary mask \(\hat{M}\) for the concept, written as \(\hat{M}=\sigma(\phi(I,s))\), together with a global confidence indicating whether the concept is present. We formalize the decoder outputs as follows: for each pixel location \(x\), the mask decoder produces a pixel-wise mask logit map \(a^{\mathrm{sam}}_{s}(x)\in\mathbb{R}\), and the pixel-wise mask probability map is obtained via sigmoid,
\begin{equation}
P^{\mathrm{sam}}_{s}(x)=\sigma\!\left(a^{\mathrm{sam}}_{s}(x)\right)\in(0,1)
\label{eq:sam3-mask-prob}
\end{equation}
In addition, it outputs an image-level presence logit \(z_{s}\in\mathbb{R}\), which characterizes the degree of concept presence and segmentability in the current image. The overall pipeline consists of two stages: (i) a unified perception encoding stage that produces dense visual representations and concept embeddings; and (ii) a mask decoding stage where the concept embedding serves as the conditioning signal and interacts cross-modally with dense visual features to yield concept-aware pixel-level responses and the final mask prediction.

\noindent\textbf{Perception Encoder (PE)}~\cite{bolya2025perception} is the unified representation module in SAM3 that maps images and text prompts into an alignable semantic space. On the image side, PE outputs a multi-level dense feature pyramid $\{F^{(l)}\}_{l=1}^{L}$ with $F^{(l)}\in\mathbb{R}^{h_l\times w_l\times d_l}$, preserving spatial layout while spanning from low-level texture cues to high-level semantic structures; on the text side, PE produces a concept embedding $e_s\in\mathbb{R}^{d}$ that serves as the conditioning signal for decoding, and can be aligned with local visual features via a similarity score $e_s^\top f_i$ to yield concept-relevant dense responses.

\section{Methodology}

The overall framework is illustrated in Fig.~\ref{fig:intro-placeholder2}. To address the unstable inter-class competition when directly applying SAM3 to open-vocabulary semantic segmentation, we extend SAM3 with a semantic-evidence branch and a unified-scale fusion mechanism. We first perform Semantic Evidence Calibration (Sec.~\ref{sec}): by matching textual concepts with intermediate-layer visual representations from the Perception Encoder, we explicitly model pixel-wise semantic--structural consistency, providing a stable semantic prior for multi-class joint inference and alleviating conflicts caused by independently generated masks. Furthermore, to reduce fluctuations induced by open-vocabulary naming, we incorporate Synonym Aggregation (Sec.~\ref{sa}): we use an LLM to expand each category into diverse linguistic variants and softly aggregate the resulting multi-view semantic evidence within the class, yielding a more robust category-level semantic response. Finally, we calibrate and fuse this semantic prior with SAM3's mask responses on a unified scale, thereby stabilizing inter-class competition, reducing overwriting and confusion, and producing more reliable open-vocabulary semantic segmentation results.

\subsection{Semantic Evidence Calibration}
\label{sec}
Semantic Evidence Calibration (SEC) aims to explicitly decouple \emph{visual separability} from \emph{semantic alignment} in multi-class open-vocabulary inference, and to calibrate the posterior under a unified additive scale, thereby stabilizing inter-class competition and reducing fluctuations induced by open-vocabulary naming. Given an input image \(I\) and a candidate concept set \(\mathcal{C}\), we first extract spatially aligned dense visual features from an intermediate layer of SAM3's perception encoder (PE), and denote the feature vector at location \(x\) as \(f(x)\in\mathbb{R}^{d}\). On the text side, for each category \(c\in\mathcal{C}\), we encode a concept embedding \(e_c\in\mathbb{R}^{d}\). Before computing similarities, both \(f(x)\) and \(e_c\) are L2-normalized so that their dot product corresponds to cosine similarity; thus, the semantic matching strength between location \(x\) and concept \(c\) is given by

\begin{equation}
u_c(x)=e_c^\top f(x)
\label{eq:sec_similarity}
\end{equation}
To align semantic evidence with SAM3's mask outputs at the same spatial resolution, we first bilinearly upsample the intermediate-layer PE feature map used to compute \(u_c(x)\) to match the resolution of the mask probability map. We then perform cross-class normalization over the candidate set \(\mathcal{C}\), converting the matching strength into an inference-ready pixel-wise semantic prior distribution:

\begin{equation}
\pi_c(x)=\frac{\exp\!\big(u_c(x)\big)}{\sum_{c'\in\mathcal{C}}\exp\!\big(u_{c'}(x)\big)}
\label{eq:semantic_prior}
\end{equation}
Here, \(\pi_c(x)\) encodes a cross-category relative preference: it does not depend on the absolute response magnitude of any single class, but instead captures which candidate concept is more semantically consistent at the same location. Therefore, when multiple concepts can yield visually plausible contours, \(\pi_c(x)\) provides a stable cross-class semantic constraint that anchors the decision to the concept with higher semantic consistency, thereby improving the stability of multi-class joint inference.

At fusion time, SEC further addresses the scale mismatch between semantic and structural evidence. Under prompt conditioning, SAM3 produces a pixel-wise response as a probability \(P^{\mathrm{sam}}_c(x)\in(0,1)\), which mainly reflects the structural segmentability of concept \(c\) at location \(x\). In contrast, \(\pi_c(x)\) is a cross-class softmax-normalized relative measure. Direct multiplication or naive comparison of the two can suffer from saturation, dominance imbalance, and incompatible numerical scales. To this end, we map the structural evidence into the logit space to obtain an additive scale, and inject the semantic prior in the log form, yielding a calibrated score under a unified scale:

\begin{figure}[!t]
  \centering
  \IfFileExists{fig/fig3.pdf}{%
    \includegraphics[width=0.99\linewidth]{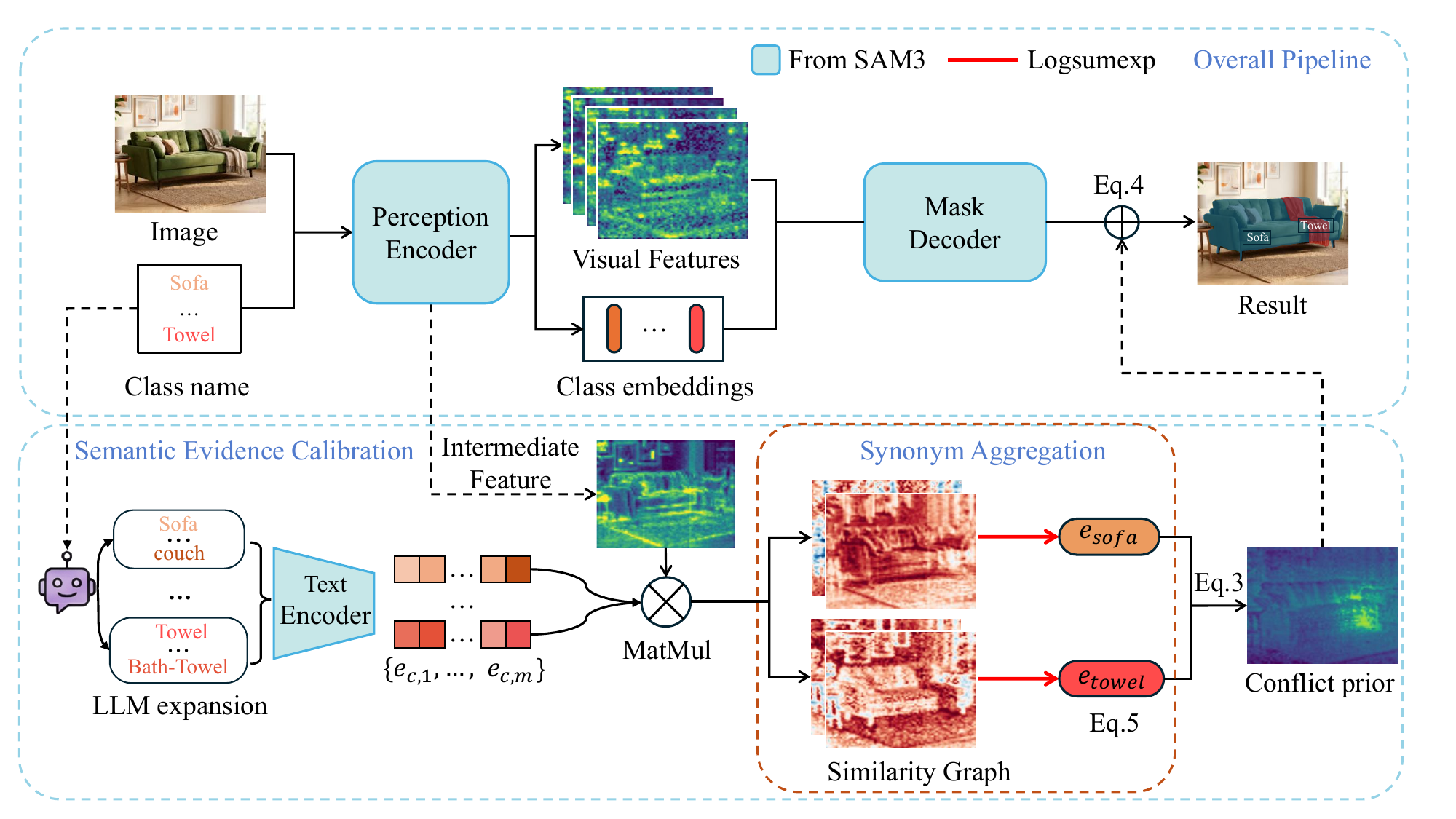}%
  }{%
    \fbox{\rule{0pt}{1.6in}\rule{0.98\linewidth}{0pt}}%
  }
  \vspace{-1.5em}
    \caption{The overview of \textbf{CoCo-SAM3}. We enhance semantic-evidence consistency via intra-class synonym aggregation and build a unified-scale conflict prior to stabilize inter-class competition, yielding stable open-vocabulary semantic segmentation.}
  \label{fig:intro-placeholder2}
\end{figure}

\begin{equation}
S_c(x)=\log\frac{P^{\mathrm{sam}}_c(x)}{1-P^{\mathrm{sam}}_c(x)}+\lambda_{\mathrm{prior}}\log \pi_c(x)+z_c
\label{eq:sec_score}
\end{equation}

The first term converts SAM3's pixel-wise structural evidence into an additive logit representation, while the second term \(\lambda_{\mathrm{prior}}\log \pi_c(x)\) explicitly injects cross-class semantic constraints as a posterior correction, enlarging inter-class margins and discouraging semantically inconsistent attribution. The last term \(z_c\) is SAM3's image-level \emph{presence logit} produced by the mask decoder; it is a class-wise scalar and is broadcast and added to all pixel locations when computing \(S_c(x)\), providing an image-level presence bias and global suppression. Intuitively, when the semantic prior favors a category, \(\log \pi_c(x)\) systematically increases inter-class separation and reduces overwriting; when the semantic distribution is flatter and more uncertain, the decision relies more on the structural logit, avoiding domination by noisy priors. The final pixel-wise prediction is \(\hat{y}(x)=\arg\max_{c\in\mathcal{C}} S_c(x)\).

\subsection{Synonym Aggregation}
\label{sa}
In multi-class open-vocabulary inference, whether inter-class conflicts can be stably suppressed often hinges on whether intra-class evidence can form a stable representative signal: if the same concept activates substantially different semantic evidence under different linguistic realizations, its relative advantage in inter-class competition will fluctuate with wording variations, thereby amplifying overwriting and label instability. To enhance intra-class consistency and reduce sensitivity to prompt wording, we introduce \emph{Synonym Aggregation} on the semantic-evidence branch, representing each category \(c\) by a set of synonymous prompts \(\mathcal{S}_c=\{s_{c,1},\ldots,s_{c,m_c}\}\). This set can be obtained by expanding the class name with an LLM to cover aliases, abbreviations, colloquial names, and dataset nomenclature, providing multi-view linguistic observations for the same concept.

Concretely, each synonym prompt \(s_{c,j}\) is encoded into a text embedding \(e_{c,j}\in\mathbb{R}^{d}\) and matched against the spatially aligned dense feature \(f(x)\) from an intermediate PE layer. The matching strength follows the definition in Sec.~\ref{sec} (see Eq.~\ref{eq:sec_similarity}), where we simply replace the category embedding \(e_c\) with the synonym embedding \(e_{c,j}\), yielding a synonym-level similarity response \(u_{c,j}(x)\). This step corresponds to the \emph{Similarity Graph} in Fig.~\ref{fig:intro-placeholder2}: different synonym prompts of the same category produce multiple similarity heatmaps.

To balance broader linguistic coverage and maintaining discriminability, we adopt a soft intra-class aggregation in the LogSumExp form and define the category-level semantic score as
\begin{equation}
\tilde u_c(x)=\log\sum_{j=1}^{m_c}\exp\!\left(\frac{e_{c,j}^{\top}f(x)}{\tau_s}\right)
\label{eq:syn_agg}
\end{equation}
The temperature \(\tau_s\) controls the sharpness of the aggregation: when \(\tau_s\) is small, \(\tilde{u}_c(x)\) behaves more like selecting the best-matching synonym (i.e., approximate max pooling), which suppresses the influence of weak or noisy synonyms; when multiple synonym prompts give consistent support at the same location, the LogSumExp aggregation accumulates multi-path evidence, enhancing the stability and robustness of category-level semantic responses. We then replace the similarity term in Sec.~\ref{sec} used to construct the semantic prior with \(\tilde{u}_c(x)\), and apply the same cross-class normalization over the candidate set \(\mathcal{C}\) to obtain \(\pi_c(x)\), which is subsequently fed into SEC for unified-scale calibration and inter-class competition.

Importantly, synonym aggregation only operates on the semantic-evidence branch: its computation occurs between the intermediate PE features and the text embeddings, and does \emph{not} incur additional SAM3 mask decoding calls. Therefore, this strategy keeps the computational overhead manageable while substantially improving intra-class consistency under prompt variation, cross-domain naming discrepancies, and long-tail concept scenarios—providing a more reliable semantic foundation for stable inter-class competition.

\section{Experiments}
\label{sec:exp}

\subsection{Experimental Setup}
\noindent{\normalsize\textbf{Datasets.}}
Following prior work, we evaluate our method on the validation sets of five datasets, which together form eight OVSS benchmarks.
Specifically, the Pascal VOC~\cite{everingham2010pascal} validation set contains 1{,}449 images and yields two benchmarks: \textbf{V21} evaluates 21 classes including the background class, while \textbf{V20} evaluates 20 classes without the background class.
The Pascal Context~\cite{mottaghi2014role} validation set contains 5{,}104 images and similarly yields two benchmarks: \textbf{PC60} evaluates 60 classes including background, while \textbf{PC59} evaluates 59 classes without background.
The COCO-Stuff~\cite{caesar2018coco} validation set contains 5{,}000 images and is annotated with 171 classes, which yields two benchmarks: \textbf{COCO-S} evaluates 171 classes without the background class; \textbf{COCO-O} is derived from COCO-Stuff by merging all stuff categories into the background class, and thus evaluates 81 classes including background.
In addition, the ADE20K~\cite{zhou2017scene} validation set contains 2{,}000 images and evaluates 150 classes without background, forming the \textbf{ADE} benchmark; the Cityscapes~\cite{cordts2016cityscapes} validation set contains 500 images and evaluates 19 classes without background, forming the \textbf{City} benchmark.
We report mean Intersection-over-Union (mIoU) as the evaluation metric.

\noindent\textbf{Experimental Details.}
All experiments uniformly adopt the official SAM3 model as the backbone and follow its default architecture and inference settings. The visual encoder uses PE-L+ as the feature backbone, and the core visual representation module is kept fully frozen throughout all experiments (i.e., no training updates are performed). The input image is resized to $1008 \times 1008$ pixels to match SAM3's native resolution. Class synonyms are expanded from the dataset-provided category texts via text-only rewriting and augmentation, without introducing any new visual categories; details are provided in the appendix. For synonym aggregation, we set the temperature to $\tau_s = 0.10$ and compute text embeddings in batches with \texttt{chunk}$=16$ to improve efficiency. We report mean Intersection-over-Union (mIoU) as the primary evaluation metric and run inference on a machine equipped with $4\times$ NVIDIA RTX 4090 GPUs (24GB each). During testing, we do not apply any data augmentation or additional post-processing.

\subsection{Comparison with State-of-the-Art Methods}

\begin{table}[t]
    \caption{Quantitative comparison with existing methods on eight OVSS benchmarks (mIoU, \%). Existing methods are grouped into four paradigms: training-based, and training-free methods based on CLIP-only, CLIP-VFM, and SAM3. \emph{Avg.} denotes the average over all eight benchmarks. \best{Red} indicates the best method, and \second{blue} indicates the second best.}
  \label{tab:ovs-natural}
  \centering
  \setlength{\tabcolsep}{1.8pt}
  \renewcommand{\arraystretch}{1.10}
  \scriptsize
  \resizebox{\textwidth}{!}{%
  \begin{tabular}{@{}lcccccccc|c@{}}
    \toprule
    \textbf{Method} & \multicolumn{3}{c}{\textit{with background}} & \multicolumn{5}{c}{\textit{without background}} & \textbf{Avg.} \\
    \cmidrule(lr){2-4}\cmidrule(lr){5-9}
    & V21 & PC60 & COCO-O & V20 & PC59 & COCO-S & City & ADE & \\
    \midrule
    \rowcolor{gray!18}\multicolumn{10}{c}{\textbf{Training-based}} \\
    GroupViT~\cite{xu2022groupvit} \venue{CVPR'22} & 50.4 & 18.7 & 27.5 & 79.7 & 23.4 & 15.3 & 11.1 & 9.2 & 29.4 \\
    TCL~\cite{ru2023token} \venue{CVPR'23} & 51.2 & 24.3 & 30.4 & 77.5 & 30.3 & 19.6 & 23.1 & 14.9 & 33.9 \\
    CLIP-DINOiser~\cite{wysoczanska2024clip} \venue{ECCV'24} & 62.1 & 32.4 & 34.8 & 80.9 & 35.9 & 24.6 & 31.7 & 20.0 & 40.3 \\
    Talk2DINO~\cite{barsellotti2025talking} \venue{ICCV'25} & 65.8 & 37.7 & 45.1 & 88.5 & 42.4 & 30.2 & 38.1 & 22.5 & 46.3 \\
    \midrule
    \rowcolor{gray!8}\multicolumn{10}{c}{\textbf{Training-free CLIP-Only}} \\
    CLIP~\cite{radford2021learning} \venue{ICML'21} & 18.6 & 7.8 & 6.5 & 49.1 & 11.2 & 7.2 & 6.7 & 3.2 & 13.8 \\
    CaR~\cite{huang2022car} \venue{CVPR'24} & 48.6 & 13.6 & 15.4 & 73.7 & 18.4 & -- & -- & 5.4 & -- \\
    CLIPtrase~\cite{shao2024explore} \venue{ECCV'24} & 50.9 & 29.9 & 43.6 & 81.0 & 33.8 & 22.8 & -- & 16.4 & -- \\
    ClearCLIP~\cite{lan2024clearclip} \venue{ECCV'24} & 51.8 & 32.6 & 33.0 & 80.9 & 35.9 & 23.9 & 30.0 & 16.7 & 38.1 \\
    SCLIP~\cite{wang2024sclip} \venue{ECCV'24} & 59.1 & 30.4 & 30.5 & 80.4 & 34.1 & 22.4 & 32.2 & 16.1 & 38.2 \\
    NACLIP~\cite{hajimiri2025pay} \venue{WACV'25} & 58.9 & 32.2 & 33.2 & 79.7 & 35.2 & 23.3 & 35.5 & 17.4 & 39.4 \\
    SFP~\cite{jin2025feature} \venue{ICCV'25} & 63.9 & 37.2 & 37.9 & 84.5 & 39.9 & 26.4 & 41.1 & 20.8 & 44.0 \\
    RF-CLIP~\cite{li2025target} \venue{AAAI'26} & 67.2 & 37.9 & 39.1 & 87.0 & 41.4 & 27.5 & 43.0 & 21.0 & 45.5 \\
    \rowcolor{gray!8}\multicolumn{10}{c}{\textbf{Training-free CLIP-VFM}} \\
    FreeDA~\cite{barsellotti2024training} \venue{CVPR'24} & 51.8 & 35.3 & 36.3 & 84.3 & 39.7 & 25.7 & 34.1 & 20.8 & 41.0 \\
    ProxyCLIP~\cite{lan2024proxyclip} \venue{ECCV'24} & 58.6 & 33.8 & 37.4 & 83.0 & 37.2 & 25.4 & 33.9 & 19.7 & 41.1 \\
    CASS~\cite{kim2025distilling} \venue{CVPR'25} & 65.8 & 36.7 & 37.8 & 87.8 & 40.2 & 26.7 & 39.4 & 20.4 & 44.4 \\
    CorrCLIP~\cite{zhang2025corrclip} \venue{ICCV'25} & 76.7 & 44.9 & 49.4 & 91.5 & 50.8 & \second{34.0} & 51.1 & 30.7 & 53.6 \\
    Trident~\cite{shi2025harnessing} \venue{ICCV'25} & 67.1 & 38.6 & 41.1 & 84.5 & 42.2 & 28.3 & 42.9 & 21.9 & 45.8 \\
    ReME~\cite{xuan2025reme} \venue{ICCV'25} & \second{82.2} & 44.6 & 48.2 & \second{93.2} & \second{53.1} & 33.3 & 59.0 & 28.2 & 55.2 \\
    \midrule
    \rowcolor{gray!8}\multicolumn{10}{c}{\textbf{Training-free SAM3}} \\
    SAM3~\cite{carion2025sam} \venue{ICLR'26} & 81.9 & \second{46.1} & \second{65.4} & 88.9 & 50.0 & 33.3 & \second{62.3} & \second{31.8} & \second{57.5} \\
    \textbf{CoCo-SAM3} (Ours) & \best{86.8} & \best{50.5} & \best{67.9} & \best{95.2} & \best{61.2} & \best{43.6} & \best{70.7} & \best{38.3} & \best{64.3} \\
    \bottomrule
  \end{tabular}
  }
\end{table}

As shown in Table~\ref{tab:ovs-natural}, our method \textbf{CoCo-SAM3} achieves the best performance across all eight OVSS benchmarks, reaching an average mIoU of \textbf{64.3}. We group the compared approaches into four paradigms: training-based OVSS methods (GroupViT~\cite{xu2022groupvit}, TCL~\cite{ru2023token}, CLIP-DINOiser~\cite{wysoczanska2024clip}, Talk2DINO~\cite{barsellotti2025talking}), training-free methods relying solely on CLIP (CLIP~\cite{radford2021learning}, CaR~\cite{huang2022car}, CLIPtrase~\cite{shao2024explore}, ClearCLIP~\cite{lan2024clearclip}, SCLIP~\cite{wang2024sclip}, NACLIP~\cite{hajimiri2025pay}, SFP~\cite{jin2025feature}, RF-CLIP~\cite{li2025target}), and training-free CLIP-VFM methods that incorporate vision foundation models (FreeDA~\cite{barsellotti2024training}, ProxyCLIP~\cite{lan2024proxyclip}, CASS~\cite{kim2025distilling}, CorrCLIP~\cite{zhang2025corrclip}, Trident~\cite{shi2025harnessing}, ReME~\cite{xuan2025reme}). Without any additional training, \textbf{CoCo-SAM3} not only significantly outperforms the vanilla SAM3~\cite{carion2025sam} framework (57.5, {\color{red}{+6.8}}), but also surpasses the strongest CLIP-VFM baselines ReME (55.2, {\color{red}{+9.1}}) and CorrCLIP (53.6, {\color{red}{+10.7}}). Further analysis by evaluation protocol shows that under the \emph{with-background} setting, \textbf{CoCo-SAM3} improves over SAM3 by about {\color{red}{+3.9}} on average, while under the more challenging \emph{no-background} setting, the average improvement increases to about {\color{red}{+8.5}}. These results indicate that our method consistently enhances class discriminability under both protocols, with more pronounced gains when the background class is absent and inter-class competition is stronger.

\begin{figure}[!t]
  \centering
  \IfFileExists{fig/fig4.pdf}{%
    \includegraphics[width=0.99\linewidth]{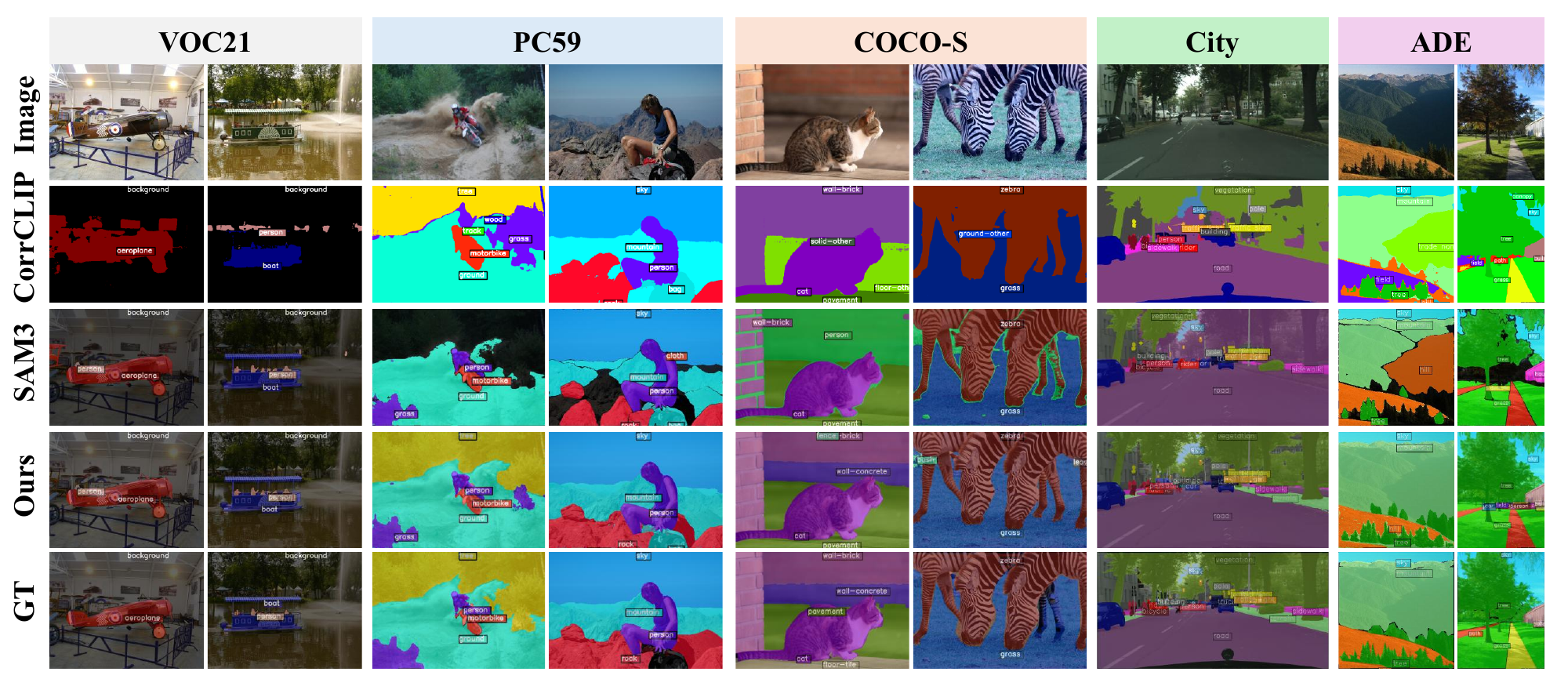}%
  }{%
    \fbox{\rule{0pt}{1.6in}\rule{0.98\linewidth}{0pt}}%
  }
  \vspace{-0.5em}
    \caption{Qualitative comparisons. ``GT'' denotes the ground truth.}
  \label{fig:intro-placeholder4}
\end{figure}

As shown in Fig.~\ref{fig:intro-placeholder4}, we conduct qualitative comparisons among CorrCLIP, SAM3, and our method on five benchmarks: VOC21, PC59, COCO-S, City, and ADE. We observe that CorrCLIP is often limited by interference among semantically similar categories, leading to incorrect category responses and local noisy predictions. Although SAM3 provides strong boundaries and fine-grained details, it still suffers from cross-class overwriting and unstable decisions in the multi-class open-vocabulary setting, which becomes more pronounced for semantically similar or co-occurring concepts. In contrast, our method performs unified pixel-wise competitive inference by jointly integrating structural and semantic evidence, thereby more robustly suppressing overwriting and confusion induced by cross-class competition while preserving SAM3's boundary advantages. As a result, our method yields more consistent category assignments over target regions, improves overall object continuity, and significantly reduces background misclassification and fragmented noise.

\subsection{Ablation Study}

\noindent\textbf{Impact of component integration.}
Table~\ref{tab:ablation-s3c-syn} evaluates the roles of the two key components in CoCo-SAM3---Semantic Evidence Calibration (SEC) and Synonym Aggregation (SA)---for open-vocabulary multi-class segmentation. Starting from the SAM3 baseline (53.3 mIoU on average), enabling SEC alone brings stable and substantial improvements, raising the average mIoU to 60.4 (\textcolor{red}{+7.1}). This gain is consistent across all benchmarks: VOC20 improves from 88.9 to 94.8, PC59 from 50.0 to 59.1, COCO-Stuff from 33.3 to 43.0, City from 62.3 to 67.9, and ADE from 31.8 to 37.3. These results highlight that calibrating multi-class evidence onto a unified and inter-class comparable scale is crucial for alleviating scale mismatch and suppressing inter-class conflicts. Further adding SA on top of SEC yields additional consistent gains, improving the average mIoU to 61.8 (\textcolor{red}{+1.4}), with more noticeable boosts on PC59 to 61.2 (\textcolor{red}{+2.1}) and City to 70.7 (\textcolor{red}{+2.8}). Overall, combining SEC and SA achieves the best performance, increasing the average mIoU from 53.3 to 61.8 (\textcolor{red}{+8.5}), demonstrating their complementarity: SEC strengthens stable inter-class competition and conflict suppression, while SA further enhances intra-class evidence consistency and integration within the same calibrated evidence space, leading to better overall performance.

\begin{table}[t]
\caption{Key component ablation study. $\checkmark$ indicates the module is enabled, and $\times$ indicates the module is removed.}
  \centering
  \vspace{-0.5em}
  \setlength{\tabcolsep}{6pt}
  \renewcommand{\arraystretch}{1.08}
  \scriptsize
  \begin{tabular}{cc|ccccc|c}
    \toprule
    \textbf{SEC} & \textbf{SA} & \textbf{VOC20} & \textbf{PC59} & \textbf{COCO-S} & \textbf{City} & \textbf{ADE} & \textbf{Avg.} \\
    \midrule
    $\times$ & $\times$ & 88.9 & 50.0 & 33.3 & 62.3 & 31.8 & 53.3 \\
    $\checkmark$ & $\times$ & 94.8 & 59.1 & 43.0 & 67.9 & 37.3 & 60.4 \\
    \rowcolor{gray!8}$\checkmark$ & $\checkmark$ & \textbf{95.2} & \textbf{61.2} & \textbf{43.6} & \textbf{70.7} & \textbf{38.3} & \textbf{61.8} \\
    \bottomrule
  \end{tabular}
  \label{tab:ablation-s3c-syn}
\end{table}

\noindent{\normalsize\textbf{Effect of Feature Layers from the Perception Encoder.}}
We further analyze how extracting features from different layers of the \emph{Perception Encoder} (PE) affects performance. PE consists of 32 stacked encoder blocks, and features at different depths exhibit a clear trade-off between semantic abstraction and the preservation of spatial details. For open-vocabulary semantic segmentation, our key goal is to better model \emph{semantic-level inter-class conflict and cooperation}: on the one hand, enhancing semantic separability to reduce confusion among categories; on the other hand, retaining sufficient local structural cues to support pixel-level boundaries and region prediction. To this end, we keep all other settings unchanged and select PE layer \#3, \#12, \#18, \#25, and \#31 as feature sources, evaluating on VOC20, PC59, COCO-S, City, and ADE. The results show that \textbf{mid-level features (layer \#18)} yield the best and most stable overall performance. This observation is consistent with the analysis in the PE paper~\cite{bolya2025perception}, where mid-level representations often strike a more favorable balance between high-level semantics and low-level visual details, making them particularly suitable for modeling category discrimination and collaboration in open-vocabulary settings.

\begin{figure}[!t]
  \centering
  \IfFileExists{fig/fig5.pdf}{%
    \includegraphics[width=0.99\linewidth]{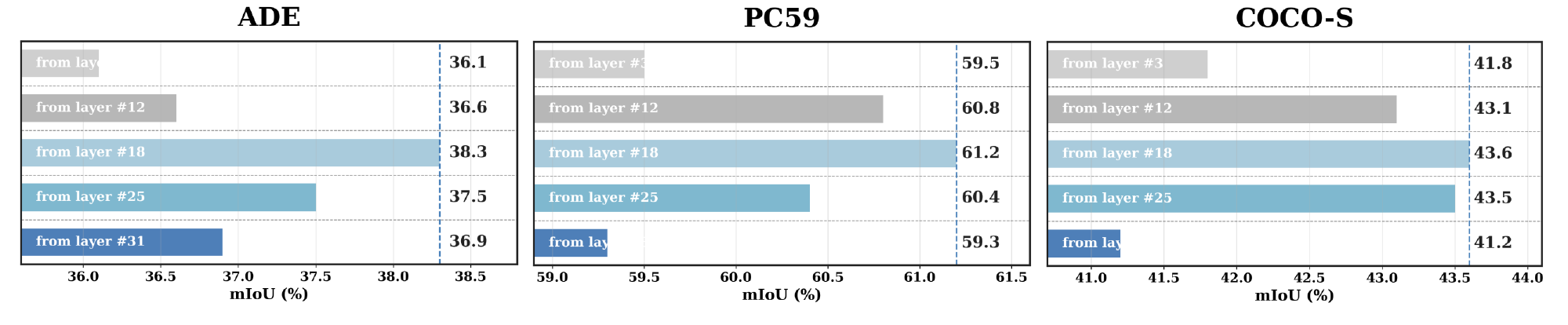}%
  }{%
    \fbox{\rule{0pt}{1.6in}\rule{0.98\linewidth}{0pt}}%
  }
  \vspace{-1em}
\caption{Comparison of mIoU using different PE layers (\#3/\#12/\#18/\#25/\#31).}
  \label{fig:intro-placeholder5}
\end{figure}

\begin{table}[t]
    \caption{Effect of selecting different feature layers from the PE on open-vocabulary semantic segmentation performance (mIoU, \%). With all other settings fixed, we extract dense features from different PE layers (\#3/\#12/\#18/\#25/\#31) }
    \vspace{-0.5em}
  \centering
  \setlength{\tabcolsep}{5pt}
  \renewcommand{\arraystretch}{1.05}
  \scriptsize
  \begin{tabular}{lcccccc}
    \toprule
    \textbf{Approach} & \textbf{VOC20} & \textbf{PC59} & \textbf{COCO-S} & \textbf{City} & \textbf{ADE}  \\
    \midrule
    \rowcolor{gray!8}\textit{layer from Perception Encoder} & & & & &  \\
    from layer \#3 & 93.2 & 59.5 & 41.8 & 69.5 & 36.1  \\
    from layer \#12 & 94.4 & 60.8 & 43.1 & 70.3 & 36.6  \\
    \rowcolor{gray!8}from layer \#18 & \textbf{95.2} & \textbf{61.2} & \textbf{43.6} & \textbf{70.7}& \textbf{38.3}  \\
    from layer \#25 & 95.0 & 60.4 & 43.5 & 70.7 & 37.5  \\
    from layer \#31 & 93.7 & 59.3 & 41.2 & 69.6 & 36.9 \\
    \bottomrule
  \end{tabular}
  \label{tab:ablation-approach2}
\end{table}



\noindent\textbf{Sensitivity to $\lambda_{\mathrm{prior}}$.}
Table~\ref{tab:ablation-approach} studies the effect of the semantic-prior weight $\lambda_{\mathrm{prior}}$ on five datasets while fixing $\tau_s=0.10$.
Overall, the method is fairly robust to $\lambda_{\mathrm{prior}}$ within a moderate range: as $\lambda_{\mathrm{prior}}$ increases from 0.3 to 0.7, performance consistently improves on PC59 (60.8$\rightarrow$61.2), COCO-S (43.1$\rightarrow$43.6), and ADE (37.8$\rightarrow$38.3), indicating that a moderately stronger cross-class semantic prior participates more effectively in competition resolution and alleviates inter-class conflicts and confusion in multi-class inference.
When further increasing $\lambda_{\mathrm{prior}}$ to 0.9, PC59 and ADE slightly drop (61.2$\rightarrow$60.7 and 38.3$\rightarrow$38.0), and VOC20 also decreases marginally (95.2$\rightarrow$94.8), suggesting that an overly strong semantic prior may suppress visual structural evidence under semantic uncertainty or among closely related categories, leading to minor degradation.
Across the five datasets, $\lambda_{\mathrm{prior}}=0.7$ achieves the best or near-best overall performance; thus it is used as the default setting in subsequent experiments.
\begin{table}[!t]
    \caption{\textbf{Sensitivity to $\lambda_{\mathrm{prior}}$ (with $\tau_s=0.10$ fixed).} mIoU (\%) on VOC20, PC59, COCO-Stuff, Cityscapes, and ADE under different $\lambda_{\mathrm{prior}}$, measuring the effect of the semantic-prior weight in logit-space fusion.}
  \centering
  \vspace{-0.5em}
  \setlength{\tabcolsep}{5pt}
  \renewcommand{\arraystretch}{1.05}
  \scriptsize
  \begin{tabular}{lccccc}
    \toprule
    \textbf{Approach} & \textbf{VOC20} & \textbf{PC59} & \textbf{COCO-S} & \textbf{City} & \textbf{ADE} \\
    \midrule
    \rowcolor{gray!8}\textit{$\tau_s = 0.10$} & & & & & \\
    $\lambda_{\text{prior}} = 0.3$ & 95.0 & 60.8 & 43.1 & 70.4 & 37.8 \\
    $\lambda_{\text{prior}} = 0.5$ & 95.2 & 61.1 & 43.3 & 70.6 & 38.1 \\
    \rowcolor{gray!8}$\lambda_{\text{prior}} = 0.7$ & \textbf{95.2} & \textbf{61.2} & \textbf{43.6} & \textbf{70.7} & \textbf{38.3} \\
    $\lambda_{\text{prior}} = 0.9$ & 94.8 & 60.7 & 43.5 & 70.7 & 38.0 \\
    \bottomrule
  \end{tabular}
  \label{tab:ablation-approach}
\end{table}
\begin{table}[t]
\centering
\caption{Efficiency comparison on OVSS benchmarks. We report the average inference time (ms/img), GPU memory consumption (MB), and segmentation performance (mIoU) averaged over eight benchmarks. chunk prompt = 16 indicates that, during multi-class inference, the evaluated concept prompts are processed in chunks of 16.}
\label{tab:efficiency}
\scriptsize
\setlength{\tabcolsep}{5pt}
\renewcommand{\arraystretch}{1.08}
\begin{tabular}{lccc}
\toprule
\textbf{Approach} & \textbf{Time(ms/img)$\downarrow$} & \textbf{Mem.(MB)$\downarrow$} & \textbf{Perf.(mIoU)$\uparrow$} \\
\midrule
\rowcolor{gray!8}\textit{chunk prompt = 16} & & & \\
SAM3 & 902 & 10563 & 57.5 \\
\textbf{CoCo-SAM3} & 934 & 10722 & 64.3 \\
\bottomrule
\end{tabular}
\end{table}

\begin{table}[t]
\centering
\caption{Ablation on prompt sources (Avg.\ mIoU). We use different LLMs to generate prompt expansions for each category and report the average mIoU under the same evaluation setting.}
\label{tab:prompt_source_ablation_avg}
\scriptsize
\setlength{\tabcolsep}{6pt}
\renewcommand{\arraystretch}{1.08}
\begin{tabular}{l c | c}
\toprule
\textbf{Variant} & \textbf{Prompt Source} & \textbf{Avg.} \\
\midrule
SAM3 (classname only) & - & 57.5 \\
\textbf{CoCo-SAM3} & Gemini-3 Pro & 64.0 \\
\rowcolor{gray!8}
\textbf{CoCo-SAM3} & GPT-5.2Thinking & 64.3 \\
\bottomrule
\end{tabular}
\end{table}
\noindent\textbf{Inference Efficiency.} Table~\ref{tab:efficiency} summarizes the runtime cost and performance during online inference. We report the average inference time (ms/img) and GPU memory consumption (MB). Notably, our method incurs no offline construction cost. \textbf{CoCo-SAM3} introduces only a minor overhead: the inference time increases from 902 ms/img to 934 ms/img, and the memory usage increases from 10563 MB to 10722 MB. This overhead mainly comes from the additional text encoding required by synonym prompt expansion and a small amount of fusion computation. Nevertheless, we fully reuse SAM3's native frozen backbone and decoding pipeline, without introducing extra models or a heavy multi-model pipeline, making the overall overhead well controlled. Meanwhile, \textbf{CoCo-SAM3} significantly improves mIoU from 57.5 to 64.3, demonstrating stable and substantial gains while maintaining inference cost close to vanilla SAM3. 

\noindent\textbf{Prompt Sources.} Table~\ref{tab:prompt_source_ablation_avg} presents an ablation study on different prompt sources. Taking SAM3 with class-name-only prompts as the baseline (57.5 mIoU), leveraging LLM-generated prompt expansions for each category and incorporating the \textbf{CoCo-SAM3} inference framework yields significant improvements: using Gemini-3 Pro achieves 64.0 mIoU (\textcolor{red}{+6.5}), while using GPT-5.2Thinking achieves 64.3 mIoU (\textcolor{red}{+6.8}). Both generators follow a consistent gain pattern, with only a 0.3 mIoU difference; These results indicate that \textbf{CoCo-SAM3} is robust to the choice of prompt generator, and GPT-5.2Thinking is slightly better, though the margin is limited.

\section{Conclusion}
We identify two key sources of instability in SAM3 for multi-class OVSS: (i) independently conditioned concept prompts make masks incomparable, causing cross-class overwriting, confusion, and unstable competition; (ii) naming diversity and synonymous expressions induce intra-class evidence drift, further increasing inconsistency. To address this, we propose a training-free method, \textbf{CoCo-SAM3}, which decomposes multi-class inference into two complementary priors---\emph{inter-class competition} and \emph{intra-class consistency enhancement}---and unifies structural and semantic evidence with a pixel-wise competitive score for stable joint prediction on a common scale. This strengthens SAM3 for multi-class OVSS without a heavy multi-model pipeline, achieving strong results on eight benchmarks.

\clearpage
%
%
\bibliographystyle{splncs04}
\bibliography{main}

\clearpage
\appendix
\renewcommand{\theHsection}{appendix.\Alph{section}}
\section*{Appendix}
\raggedbottom
\setlength{\parskip}{0pt}
\setlength{\intextsep}{6pt}
\setlength{\textfloatsep}{6pt}
\setlength{\floatsep}{6pt}

\section{Implementation Details}
\label{sec:supp-impl}

For each category, we construct a prompt set consisting of the canonical class name and several synonymous expressions. The synonym set is used only to build the semantic prior, while the structural branch performs decoding only once using the canonical prompt. Therefore, our method does not run multiple structural predictions for multiple synonymous prompts.

Let $\mathcal{S}_c$ denote the synonym set of category $c$, and let $u_{c,s}(x)$ denote the dense matching score between prompt $s \in \mathcal{S}_c$ and the image feature at pixel $x$. We first aggregate intra-class semantic evidence as
\begin{equation}
z_c(x)=\log \sum_{s\in\mathcal{S}_c}\exp\big(u_{c,s}(x)\big),
\end{equation}
and then perform cross-class normalization over the full candidate set $\mathcal{C}$:
\begin{equation}
\log \pi_c(x)= z_c(x)-\log \sum_{c'\in\mathcal{C}}\exp\big(z_{c'}(x)\big).
\end{equation}
Thus, the semantic prior is comparatively calibrated within a given candidate set. All main results use full-class competition; the partial-negative setting in Fig.~1 is used only for analysis.

The structural branch provides spatial evidence and is combined with the semantic prior to form the final category score $Z_c(x)$. Category-level existence information acts as an image-level bias that suppresses unlikely classes. The final prediction is obtained as
\begin{equation}
\hat{y}(x)=\arg\max_{c\in\mathcal{C}} Z_c(x).
\end{equation}
For the with-background setting, background is not treated as an explicit text prompt category; instead, it is obtained by rejecting low-confidence pixels in the unified score space. In implementation, log-domain operations are computed in numerically stable forms. Compared with conventional prompt ensembling, our method uses synonym expansion only to build a competitive semantic prior, rather than averaging multiple structural predictions or final outputs.

\section{Additional Quantitative Results and Ablations}

We further provide two additional ablation studies to examine the sensitivity of our method to the source of dense visual features and the design of intra-class aggregation. Unless otherwise specified, all other settings remain the same as those in the main paper. Here we report results on five datasets: V20, PC59, COCO-S, City, and ADE.

\noindent\textbf{Replacing PE Features with DINOv2~\cite{oquab2023dinov2}.}
We replace the PE dense features in the semantic-prior branch with DINOv2 features to examine how dependent our framework is on the choice of visual representation. In this experiment, all other parts of the inference pipeline remain unchanged, including the synonym sets, cross-class normalization, and unified fusion with structural responses. The results are shown in Table~\ref{tab:dinov2_pe}. Even with DINOv2 features, the overall performance drops only moderately, indicating that the proposed competitive semantic prior is compatible with different dense feature sources rather than being tied to one specific representation. Meanwhile, PE still performs better, suggesting that a feature space more aligned with the structural branch leads to more stable semantic-structural cooperation and thus better final predictions.

\begin{table}[t]
\centering
\caption{Ablation on semantic-prior feature sources.}
\label{tab:dinov2_pe}
\setlength{\tabcolsep}{5pt}
\renewcommand{\arraystretch}{1.05}
\scriptsize
\begin{tabular}{lccccc|c}
\toprule
Method & V20 & PC59 & COCO-S & City & ADE & Avg. \\
\midrule
Ours (DINOv2)  & 94.1 & 59.8 & 42.1 & 69.2 & 36.9 & 60.4 \\
\rowcolor{gray!8}Ours (PE)      & 95.2 & 61.2 & 43.6 & 70.7 & 38.3 & 61.8 \\
\bottomrule
\end{tabular}
\end{table}

\noindent\textbf{Ablation on Intra-class Aggregation.}
We further compare three intra-class aggregation strategies, i.e., Average, Max, and log-sum-exp, to clarify whether the gain comes merely from introducing more synonymous prompts or from a more effective way of integrating semantic evidence. The results are shown in Table~\ref{tab:agg_ablation}. Average benefits from multiple prompts, but is prone to over-smoothing weakly related expressions. Max relies only on the strongest response, which makes it more selective but less robust overall. In contrast, log-sum-exp preserves the dominance of highly relevant prompts while still accumulating complementary evidence from other related expressions, leading to the best performance. This result suggests that the advantage of our method comes not only from expanding the prompt set, but also from using an aggregation strategy that better matches the competitive semantic-prior design.

\begin{table}[t]
\centering
\caption{Ablation on intra-class aggregation strategies.}
\label{tab:agg_ablation}
\setlength{\tabcolsep}{5pt}
\renewcommand{\arraystretch}{1.05}
\scriptsize
\begin{tabular}{lccccc|c}
\toprule
Aggregation & V20 & PC59 & COCO-S & City & ADE & Avg. \\
\midrule
Average      & 94.8 & 60.5 & 42.9 & 69.9 & 37.6 & 61.1 \\
Max          & 93.5 & 58.4 & 40.2 & 67.6 & 36.7 & 59.3 \\
\rowcolor{gray!8} Log-Sum-Exp  & 95.2 & 61.2 & 43.6 & 70.7 & 38.3 & 61.8 \\
\bottomrule
\end{tabular}
\end{table}

\section{Synonym Generation Template}

Inspired by ConceptBank~\cite{pei2026taming}, we generate a small set of lexical variants for each category in an offline stage. The expanded prompts are used only in the semantic-prior branch, while the structural branch always uses the canonical class name.

\noindent\textbf{Quality control.}
We keep only label-faithful variants, such as aliases, spelling variants, singular and plural variants, and hyphenation or spacing variants. We exclude hypernyms, hyponyms, parts, attributes, actions, scenes, materials, and other expressions that may change the original class boundary. The generated prompts are further lowercased, deduplicated, and filtered by length and validity.

\begin{quote}
\textbf{[System]} Generate a small set of lexical variants for a segmentation class name while strictly preserving its original visual semantics.

\textbf{[User]} Given Dataset: \{DATASET\} and Class: \{CLASS\}, output a single-line, comma-separated prompt list. The first item must be the original class name.

\textbf{Rules:} allow aliases, spelling variants, singular and plural variants, and hyphenation or spacing variants; do not generate hypernyms, hyponyms, parts, attributes, actions, scenes, materials, or other expressions that expand the original category boundary; keep each phrase concise; if no reliable variant exists, keep only the original class name; retain at most \(M=10\) candidate items for each class, including the original class name.
\end{quote}

\noindent\textbf{Prompt file and usage.}
The final prompt file is stored in plain-text format, with one category per line and comma-separated retained variants. All experiments use fixed prompt files. At inference time, the full prompt set is used only for intra-class aggregation in the semantic-prior branch.

\section{More Qualitative Results}
\label{sec:supp-qual}

\begin{figure}[H]
  \centering
  \IfFileExists{figap/pc59.pdf}{%
    \includegraphics[width=\linewidth]{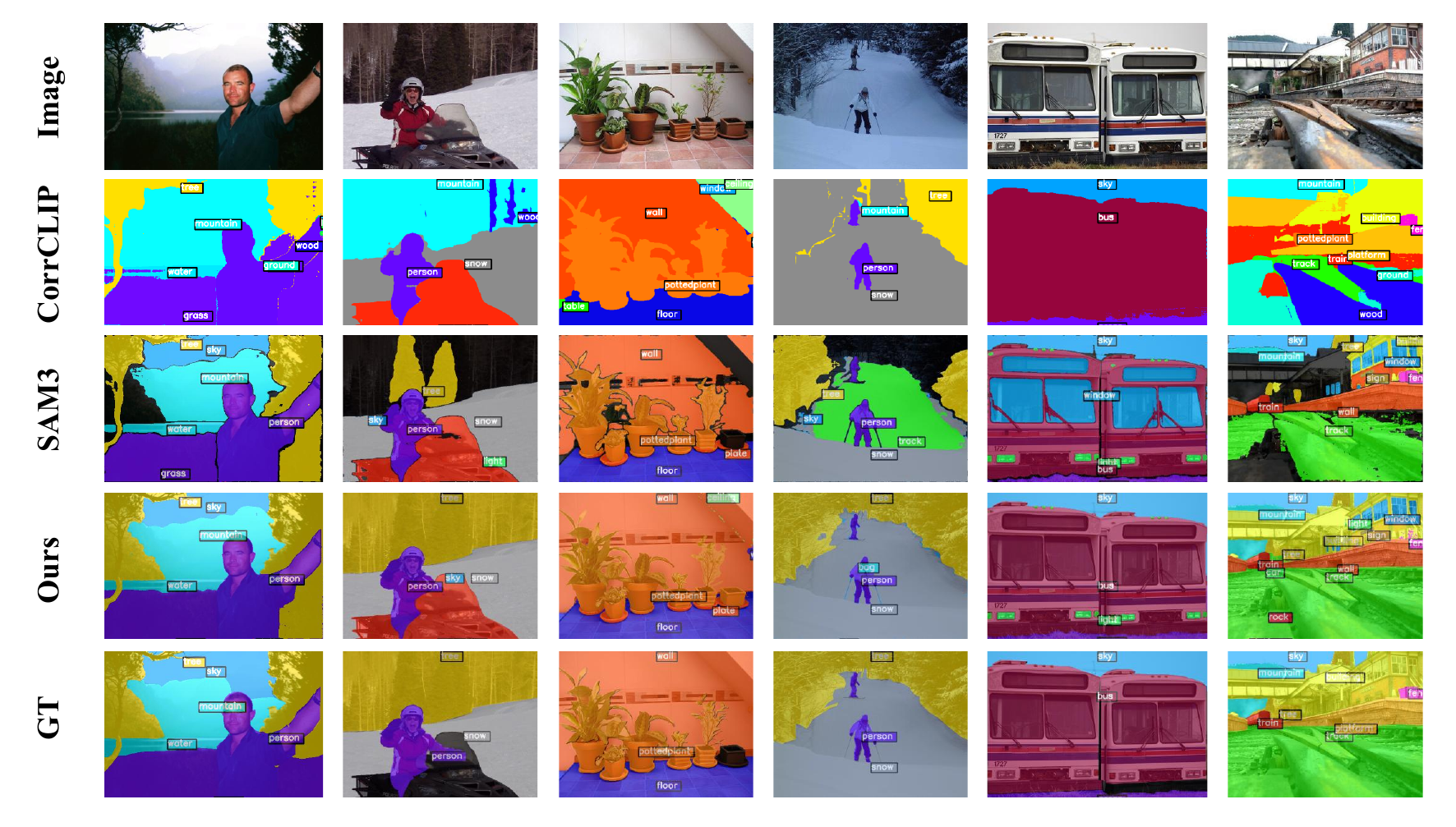}%
  }{%
    \fbox{\rule{0pt}{1.6in}\rule{0.98\linewidth}{0pt}}%
  }
  \caption{Additional qualitative results of our method on PC59.}
  \label{fig:supp-pc59}
\end{figure}

We provide more qualitative comparisons on PC59, COCO-S, VOC21, Cityscapes, and ADE20K to further demonstrate the behavior of our method under diverse scenes and category sets. Compared with CorrCLIP and SAM3, our method produces more coherent semantic regions, better preserves object boundaries, and reduces cross-class overwriting in challenging open-vocabulary settings. The corresponding results are shown in Fig.~\hyperref[fig:supp-pc59]{5--9}. These examples consistently show that the proposed competitive semantic prior, together with unified fusion with structural responses, leads to more stable pixel-wise predictions across different datasets.

\clearpage
\flushbottom
\vspace*{\fill}

\vspace{6pt}

\begin{figure}[H]
  \centering
  \IfFileExists{figap/coco-s.pdf}{%
    \includegraphics[width=\linewidth]{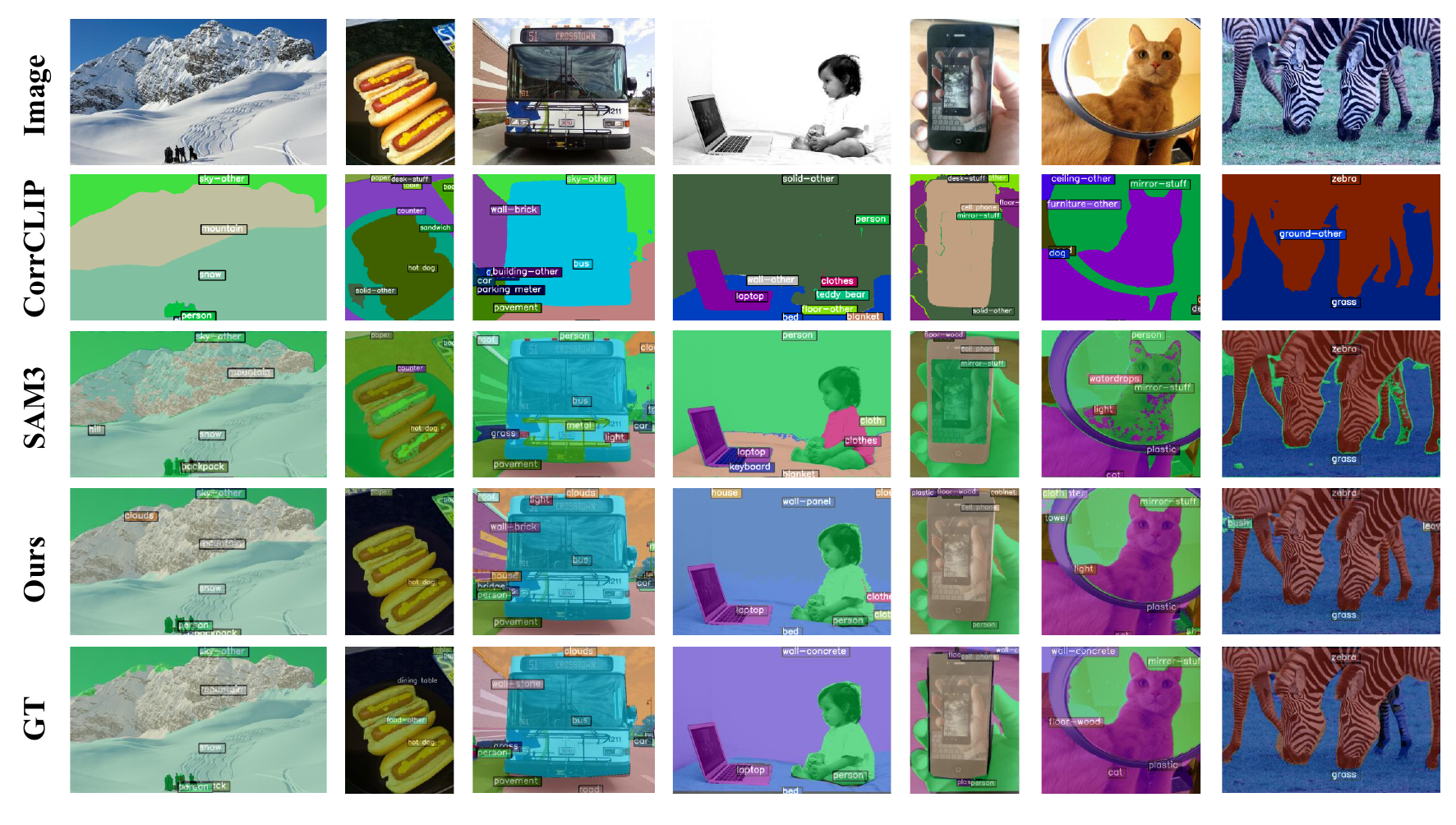}%
  }{%
    \fbox{\rule{0pt}{1.6in}\rule{0.98\linewidth}{0pt}}%
  }
  \caption{Additional qualitative results of our method on COCO-S.}
  \label{fig:supp-cocos}
\end{figure}

\vspace*{\fill}
\clearpage
\vspace*{\fill}

\begin{figure}[H]
  \centering
  \IfFileExists{figap/v21.pdf}{%
    \includegraphics[width=\linewidth]{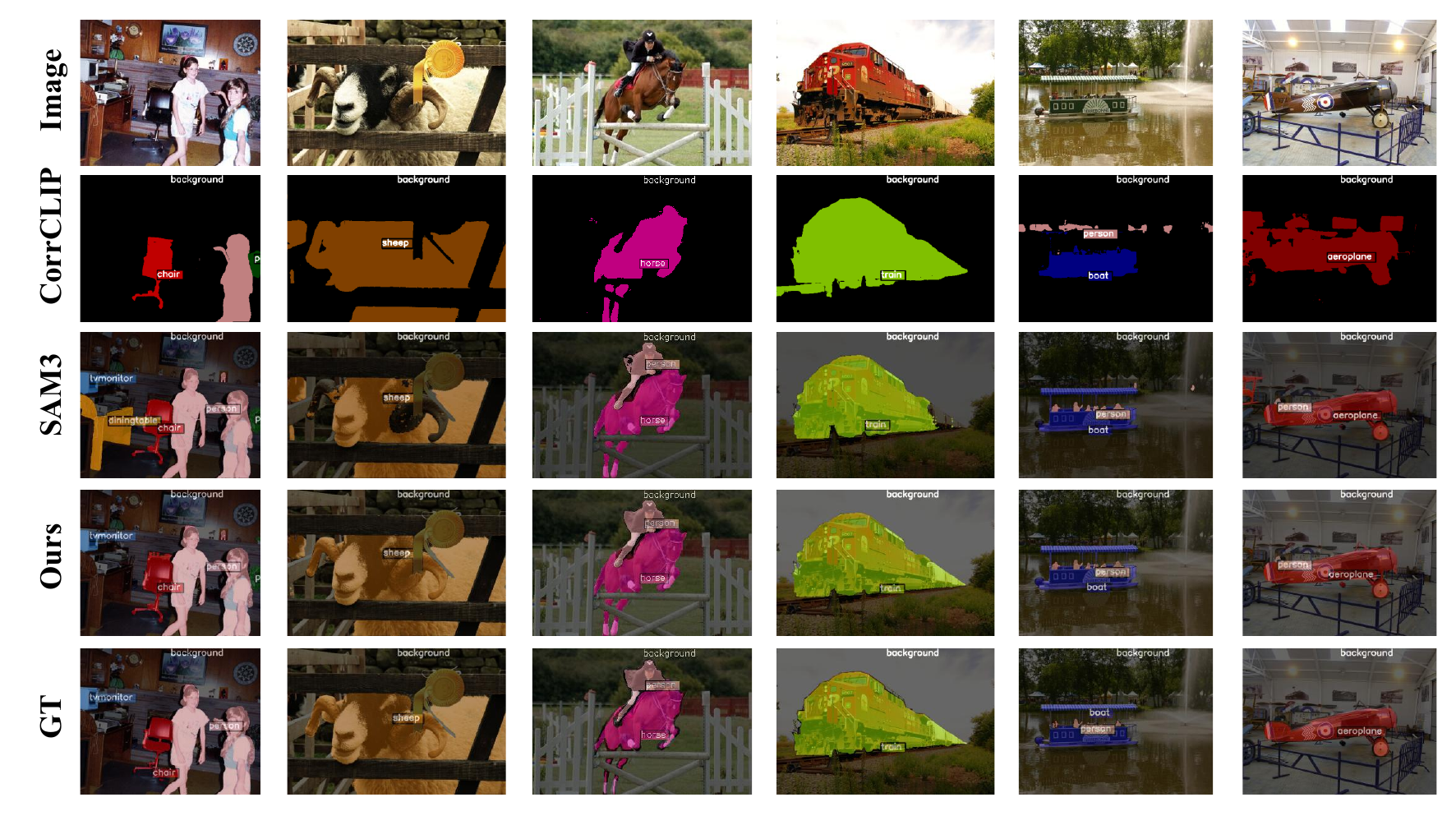}%
  }{%
    \fbox{\rule{0pt}{1.6in}\rule{0.98\linewidth}{0pt}}%
  }
  \caption{Additional qualitative results of our method on VOC21.}
  \label{fig:supp-v21}
\end{figure}

\vspace{6pt}

\begin{figure}[H]
  \centering
  \IfFileExists{figap/city.pdf}{%
    \includegraphics[width=\linewidth]{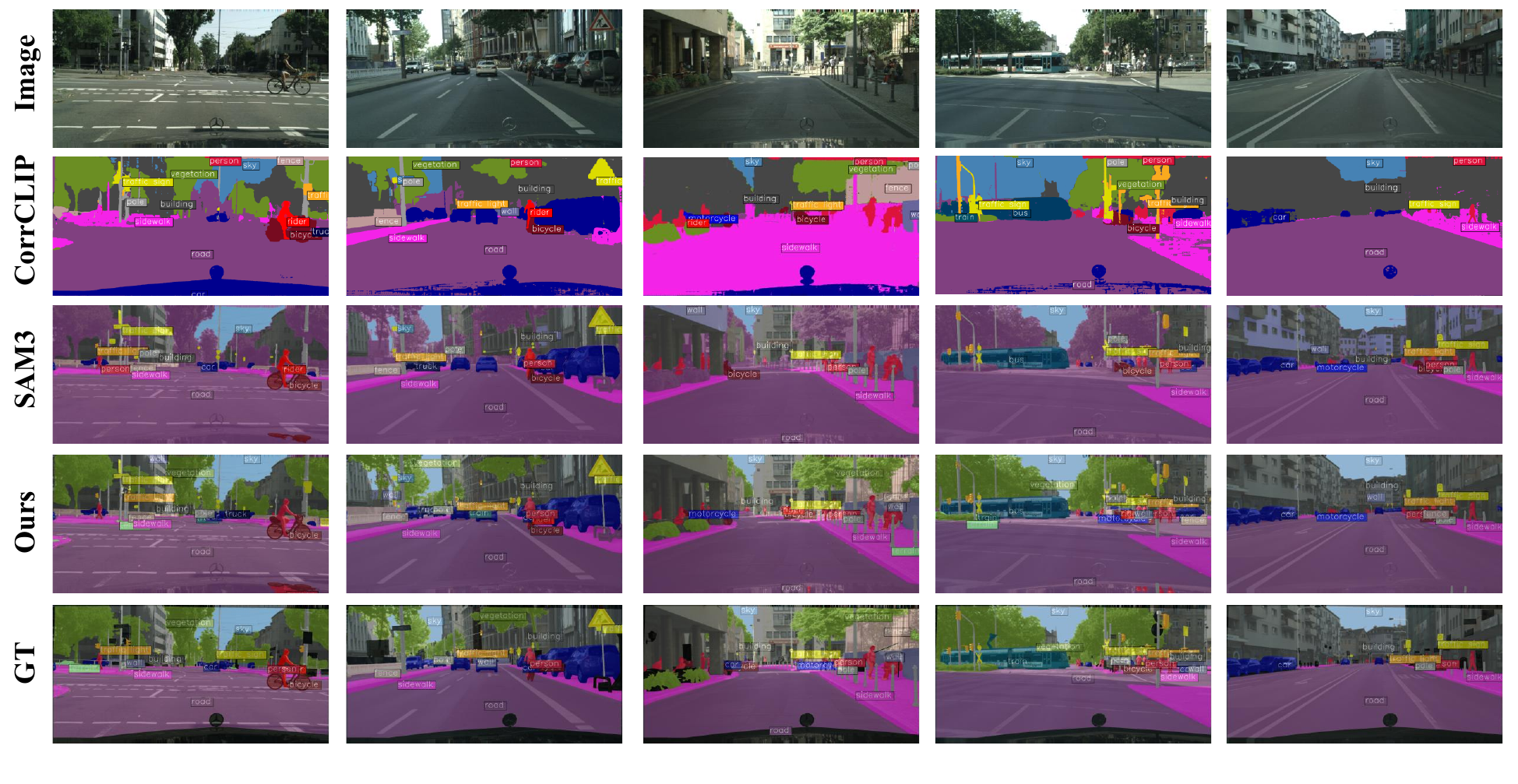}%
  }{%
    \fbox{\rule{0pt}{1.6in}\rule{0.98\linewidth}{0pt}}%
  }
  \caption{Additional qualitative results of our method on Cityscapes.}
  \label{fig:supp-city}
\end{figure}

\vspace*{\fill}
\clearpage
\vspace*{\fill}

\begin{figure}[H]
  \centering
  \IfFileExists{figap/ade.pdf}{%
    \includegraphics[width=\linewidth]{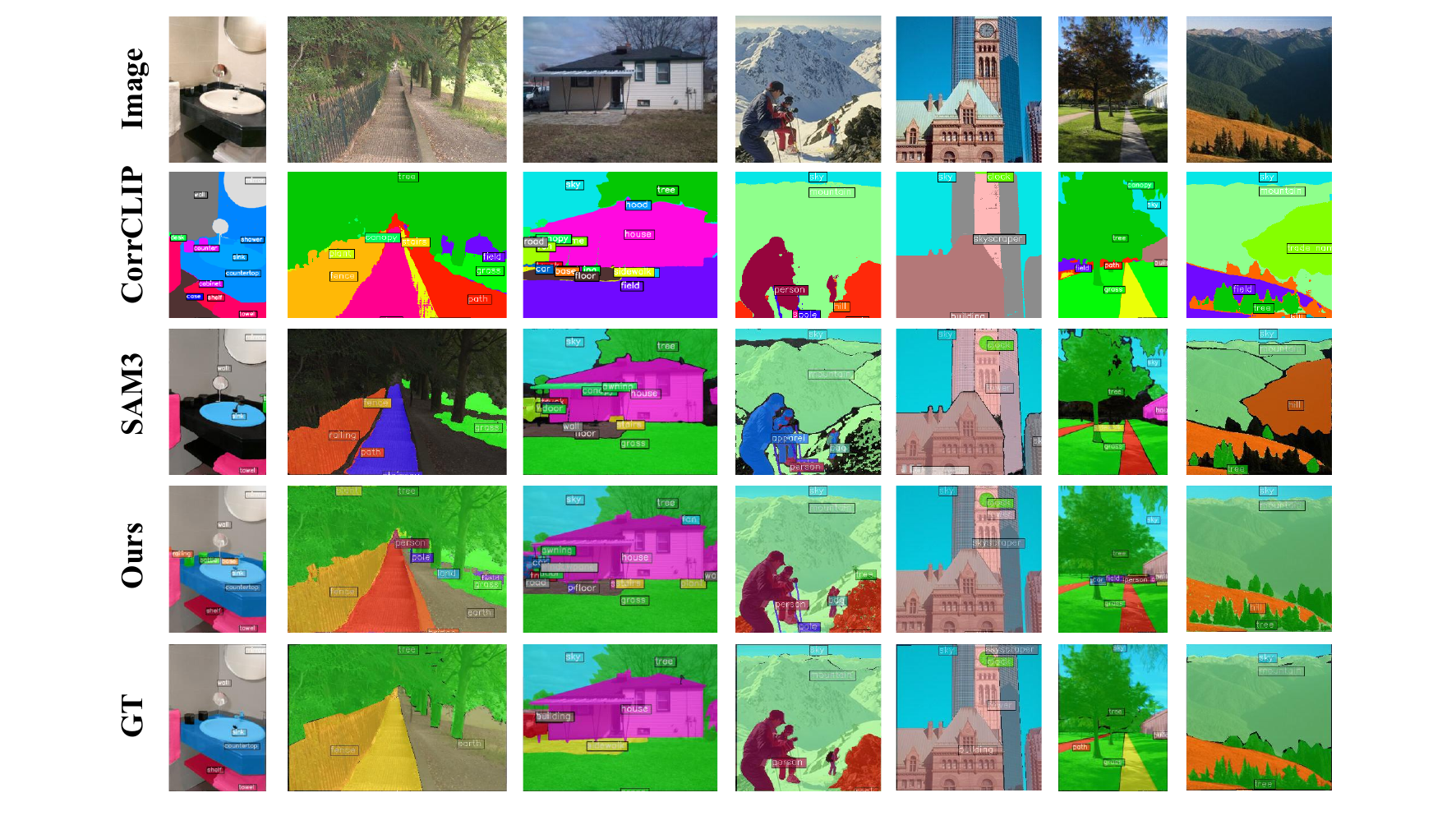}%
  }{%
    \fbox{\rule{0pt}{1.6in}\rule{0.98\linewidth}{0pt}}%
  }
  \caption{Additional qualitative results of our method on ADE20K.}
  \label{fig:supp-ade}
\end{figure}

\vspace*{\fill}

\end{document}